\documentclass[review]{elsarticle}
\usepackage{lineno,hyperref}
\usepackage{amsmath}
\usepackage{amsthm}
\usepackage{amsfonts}
\usepackage{multirow}
\usepackage{booktabs} 
\usepackage{subfigure}
\usepackage{graphicx}

\journal{Journal of \LaTeX\ Templates}

\begin{document}

\begin{frontmatter}

\title{Cross-modal Image Retrieval with Deep Mutual Information Maximization}
\author[mymainaddress,mythirdaddress]{Chunbin Gu}
\ead{guchunbin@zju.edu.cn}
\author[mymainaddress,mythirdaddress,myforthaddress]{Jiajun Bu\corref{mycorrespondingauthor}}
\ead{bjj@zju.edu.cn}
\author[mymainaddress,mythirdaddress]{Xixi Zhou}
\ead{xixi.zxx@zju.edu.cn}
\author[mymainaddress,mythirdaddress]{Chengwei Yao}
\ead{yaochw@zju.edu.cn}
\author[mysecondaryaddress,mysixthaddress]{Dongfang Ma}
\ead{mdf2004@zju.edu.cn}
\author[mymainaddress,mythirdaddress]{Zhi Yu}
\ead{yuzhirenzhe@zju.edu.cn}
\author[myfifthaddress]{Xifeng Yan}
\ead{xyan@cs.ucsb.edu}

\cortext[mycorrespondingauthor]{Corresponding author}

\address[mymainaddress]{Zhejiang Provincial Key Laboratory of Service Robot, College of Computer Science, Zhejiang University, 310007, Hangzhou, P.R.China}
\address[mysecondaryaddress]{Institute of Marine Sensing and Networking, Zhejiang University, 310058, Hangzhou, P.R.China}
\address[mythirdaddress]{Alibaba-Zhejiang University Joint Institute of Frontier Technologies, 310007, Hangzhou, P.R.China}
\address[myforthaddress]{MOE Key Laboratory of Machine Perception,100871, Beijing, P.R.China}
\address[mysixthaddress]{Key Laboratory of Ocean Observation-Imaging Testbed of Zhejiang Province, Zhejiang University, Zhoushan, 316021, P.R.China}
\address[myfifthaddress]{Department of Computer Science, University of
California, Santa Barbara, CA, 93106}


\begin{abstract}
In this paper, we study the cross-modal image retrieval, where the inputs contain a source image plus some text that describes certain modifications to this image and the desired image. Prior work usually uses a three-stage strategy to tackle this task: 1) extract the features of the inputs; 2) fuse the feature of the source image and its modified text to obtain fusion feature; 3) learn a similarity metric between the desired image and the source image + modified text by using deep metric learning. Since classical image/text encoders can learn the useful representation and common pair-based loss functions of distance metric learning are enough for cross-modal retrieval, people usually improve retrieval accuracy by designing new fusion networks. However, these methods do not successfully handle the modality gap caused by the inconsistent distribution and representation of the features of different modalities, which greatly influences the feature fusion and the similarity learning. To alleviate this problem, we adopt the contrastive self-supervised learning method Deep InforMax (DIM) \cite{hjelm2018learning} to our approach to bridge this gap by enhancing the dependence between the text, the image, and their fusion. Specifically, our method narrows the modality gap between the text modality and the image modality by maximizing mutual information between their not exactly semantically identical representation. Moreover, we seek an effective common subspace for the semantically same fusion feature and desired image's feature by utilizing Deep InforMax between the low-level layer of the image encoder and the high-level layer of the fusion network. Extensive experiments on three large-scale benchmark datasets show that we have bridged the modality gap between different modalities and achieve state-of-the-art retrieval performance.
\end{abstract}


\end{frontmatter}


\section{Introduction}
\begin{figure}
\centering
\includegraphics[scale=0.42]{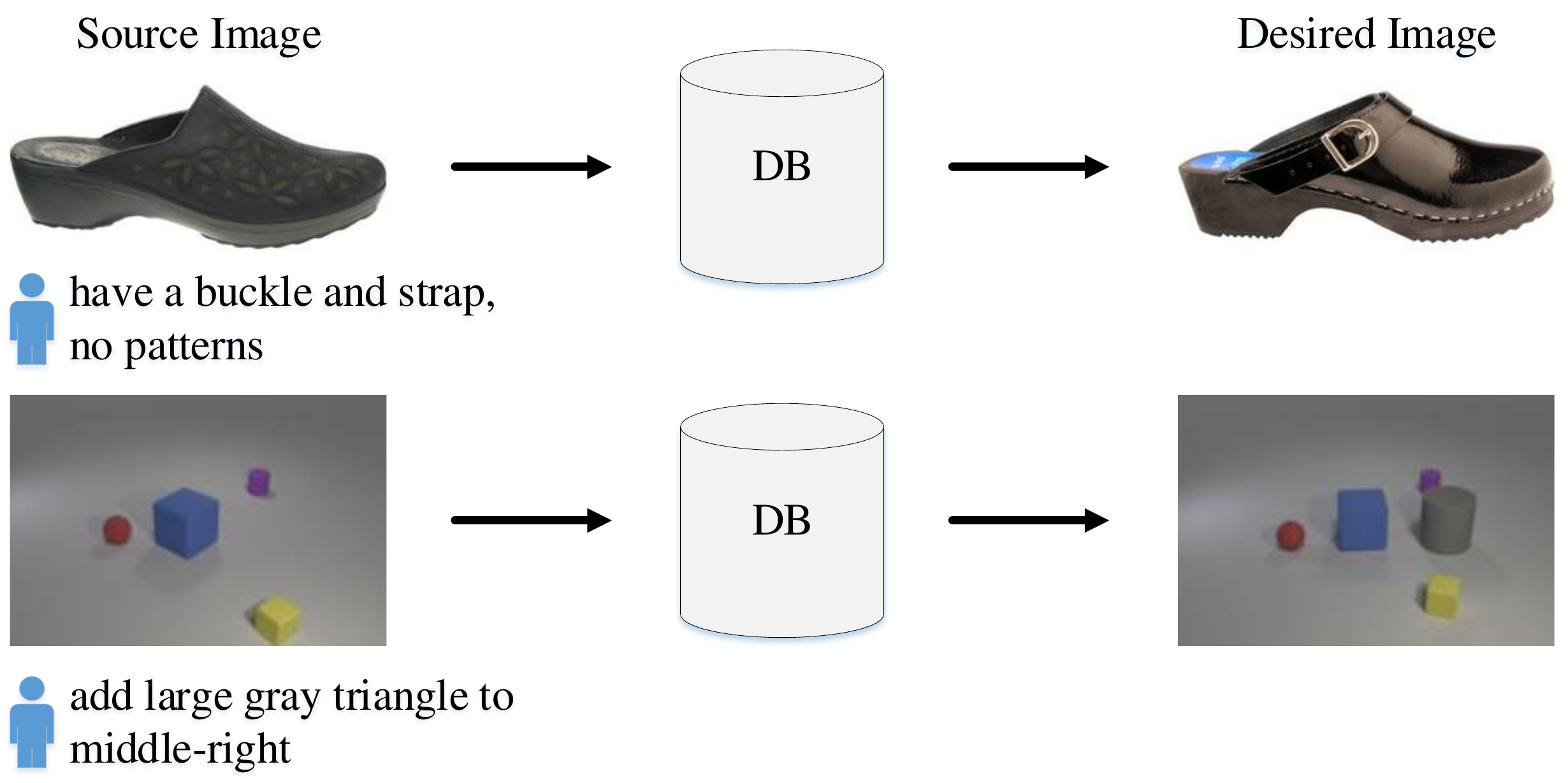}
\caption{Example of image retrieval based on image and text fusion. The text states the desired modification to the image and the information of the two input modalities conveys into the system.}
\label{fig1}
\end{figure}

Image retrieval is a key compute vision problem and it has made great progress due to deep learning \cite{chopra2005learning,gordo2016deep,cakir2019deep,cao2019hybrid,tian2020bootstrap}. Cross-modal image retrieval allows using other types of query, such as text to image retrieval \cite{wang2016learning,hu2019multi}, sketch to image retrieval \cite{sangkloy2016sketchy,pang2019generalising} and cross-view image retrieval \cite{lin2015learning,hu2018cvm}. In this paper, we consider the case where input queries are formulated as an input image plus the text that describes desired modifications to the image. Different from attribute-based image retrieval \cite{zhao2017memory}, our input text can be multi-word instead of a single attribute. For instance, our input image is a women clog and the text could be ``have a buckle and strap, no patterns". Our desired image should meet the requirement of the two input modalities (Figure \ref{fig1}).
 
To solve this problem, TIRG \cite{vo2019composing} first extracts the features of the source image and the modification text by ResNet-18 and LSTM respectively, then fuses them via a gated residual connection, and finally learns the similarity metric using softmax cross-entropy loss to minimize the distance between fusion feature and the feature of the desired image. Similarly, another research work \cite{guo2018dialog} gets the feature of the source image and that of the text by ResNet-101 and GRU, fuses them through concatenation and improves the rank of the desired image by deep metric learning based on reinforcement learning. Although there are many other related work \cite{socher2014grounded,nagarajan2018attributes,noh2016image,perez2018film}, most of them focus on designing new feature fusion networks and employing different DML loss function. These methods align the feature vector and assess the similarity between the desired image and the fusion of the source image and the modified text by common retrieval losses. However, since features of different modalities usually have inconsistent distribution and representation, a modality gap exists so as to affect the retrieval performance significantly \cite{wang2017adversarial}.

Mutual information (MI) can capture non-linear statistical dependencies between random variables and act as a measure of true dependence \cite{kinney2014equitability}. The recent research \cite{belghazi2018mine,bachman2019learning,hjelm2018learning} offers various general-purpose parametric neural estimators of mutual information between different representations in the deep neural network. Thus, we align the feature distributions of text, image, and their fusion by Deep InfoMax (DIM) \cite{hjelm2018learning} between the representations in the encoders of these modalities. Specifically, we maximize the MI between the low-level representation in the text encoder and the high-level representation in the desired image encoder (ITDIM) to projects these two modalities into a common subspace. As the text and the desired image are not exactly semantically same, we realize ITDIM by estimating their overlapping semantic information. Compared with two modalities with independent distribution (like text and images), the image modality is the key component of the fusion modality, so their features' distribution aligns with each other to some extent. Our method gets a better alignment by maximizing the MI between the low-level representation of the desired image and the fusion's high-level representation (IFDIM). Here the semantic information of the different-level representations is identical. A handful of literature in the text to image retrieval field narrows the modality gap using adversarial loss \cite{wang2017adversarial,wang2019learning}, which attempts to make the features of different modalities indistinguishable. In essence, these methods can be treated as the special cases of MINE (the basic version of Deep InfoMax) \cite{belghazi2018mine}, which maximize the MI between the last layers of different encoders using minimax objective \cite{belghazi2018mine,hjelm2018learning,nowozin2016f}. Maximizing mutual information between two complete feature vectors is often unsufficient for increasing the dependence of two modalities. Therefore, DIM also maximizes the average MI between the high-level representation and local regions of the low-level representation (e.g., patches rather than the complete image) to make the alignment better. Because our method uses the Text Image Residual Gating (TIRG) \cite{vo2019composing} as our basic network architecture, we call it TIRG-DIM.
 
The experiment shows that the proposed method can achieve higher retrieval accuracy compared to existing methods on three standard benchmark datasets, namely Fasion200K \cite{han2017automatic}, MIT-states \cite{isola2015discovering} and CSS \cite{vo2019composing}. 

To summarize, our contributions are threefold:

\begin{itemize}
    \item We design a novel framework for cross-modal image retrieval based on Deep InfoMax. By using ITDIM, maximizing MI by estimating the overlapping semantic information between the representations of the text modality and the image modality, we project the features of these two semantically different modalities with independent distribution into a common subspace, which can improve retrieval accuracy by learning a higher quality fusion feature.
    
    \item We accurately align the distribution of the features of the fusion modality and its main component, the image modality, by IFDIM that maximizes mutual information between the semantically same representations in the fusion network and the desired image encoder, which leads to more competitive retrieval results.
    
    \item The empirical results show that our method outperforms the state-of-the-art approaches for cross-modal image retrieval on three public benchmarks, Fashion200K, MIT-states and CSS.
\end{itemize}

\section{Related Work}
In this section, we briefly review the methods of cross-modal image retrieval based on feature fusion and concisely introduce deep mutual information maximization.
\subsection{Cross-modal Image Retrieval Based on Feature Fusion}
In addition to the methods mentioned before, the previous work \cite{vo2019composing} also provides seven benchmarks which use the same system pipeline as TIRG except feature fusion modules. For similarity, we define the feature of the source image, that of the modified text and the fusion feature as $\phi_s$, $\phi_t$ and $\phi_{st}$. The feature fusion methods of these benchmarks are as follows, 
\begin{itemize}
\item Image Only: $\phi_{st} = \phi_s$.
\item Text Only: $\phi_{st} = \phi_t$. 
\item Concatenating features of image and text using $\phi_{st} = f_{MLP}([\phi_s,\phi_t])$ \cite{socher2014grounded,antol2015vqa}. In experiments, it is implemented by making use of two layers of MLP with RELU, the batch-norm and the dropout rate of 0.1.
\item Show and Tell \cite{vinyals2015show}: In this method, $\phi_{st}$ is the final state of a LSTM which encoders the image and the words in the text in turn.
\item Attribute as operator: Embed each text as a transformation and apply it to $\phi_s$ to obtain $\phi_{st}$ \cite{nagarajan2018attributes}
\item Parameter Hashing \cite{noh2016image}: $\phi_{st}$ is the output of the image CNN which replaces the weights of a fc layer with transformation matrix, i.e. the hash of $\phi_t$.
\item Relationship \cite{santoro2017simple} first constructs relationship features by concatenating text feature $\phi_t$ and the feature-map vectors from the convolved image; then these features pass through a MLP and the result is averaged to get $\phi_{st}$.
\item FiLM \cite{perez2018film} outputs $\phi_{st}$ by a feature-wise affine transformation of the image feature, $\phi_{st} = \gamma_i \phi_s + \beta_i$, where $\gamma_i,\beta_i \in R^C$ is the modulation features predicted by $\phi_t$, the $i$ is the index of the layer and $C$ is the number of features or feature maps.
\end{itemize}

The above approaches learn the text feature, the desired image feature, and the fusion feature separately. This leads to the modality gap due to the inconsistent distribution of these features, which greatly affects the retrieval accuracy. To alleviate this problem, we align these distributions by maximizing the mutual information between the representations of different modalities.   
 
\subsection{Deep Mutual Information Maximization}
Mutual information (MI) is a fundamental quantity across data science for measuring the relationship between random variables \cite{becker1992information,becker1996mutual,wiskott2002slow}. Unlike correlation, MI captures non-linear statistical dependence between variables and thus can act as a measure of true dependence \cite{kinney2014equitability}. Though the infomax principle that the idea of maximizing MI between the input and output has been proposed in many traditional feature learning methods \cite{bell1995information,linsker1988self}, MI is often hard to compute \cite{paninski2003estimation}, especially for the high-dimensional and continuous variables in the deep neural network.

Fortunately, the recent research makes a theoretical breakthrough in deep mutual information estimation and provides the method for computing and optimizing the MI between input and output in a deep neural network. Mutual Information Neural Estimation (MINE) \cite{belghazi2018mine} is the first general-purpose estimator of the MI of continuous variables. Furthermore, Deep InfoMax \cite{hjelm2018learning} leverages local structure apart from global MI utilized in MINE to improve the suitability of representations for classification and provides various MI estimators. Moreover, mutual information maximization between features extracted from multiple views also draws much attention \cite{bachman2019learning,tian2019contrastive}, and these studies demonstrate that the quality of the representation improves as the number of views increases. As a member of self-supervised learning \cite{oord2018representation,henaff2019data,he2019momentum,chen2020simple}, deep mutual information maximization exploiting dual optimization to estimate divergences goes beyond the minimax objective as formalized in GANs \cite{goodfellow2014generative,arjovsky2017towards,arjovsky2017wasserstein}. Many deep learning tasks have adopted this method for estimating MI via back-propagation and proven its effectiveness, like text generation \cite{zhang2018generating,qian2019enhancing,mccarthy2019improved} and representation learning \cite{kong2019mutual,tschannen2019mutual,wen2020mutual}. 

In the cross-modal field, the use of deep mutual information is diverse. Since the mutual information between different modalities usually has higher semantic meaning compared to information that is modality-specific, people verify whether two input data correspond to each other by capturing mutual information between the two modalities \cite{sayed2018cross,jing2020self}. Also, the researchers utilize mutual information estimation to improve the qualities of representations \cite{guo2019learning,vemulapalli2017deep}. In the visual question answering field, Information Maximization Visual Question Generator \cite{krishna2019information} employs mutual information maximization to guarantee relevance between the generated question with the image and the expected answer. To our best knowledge, there are few research utilizing deep mutual information maximization to align the features' distributions from different modalities up till now.

\begin{figure}
\centering
\includegraphics[scale=0.3]{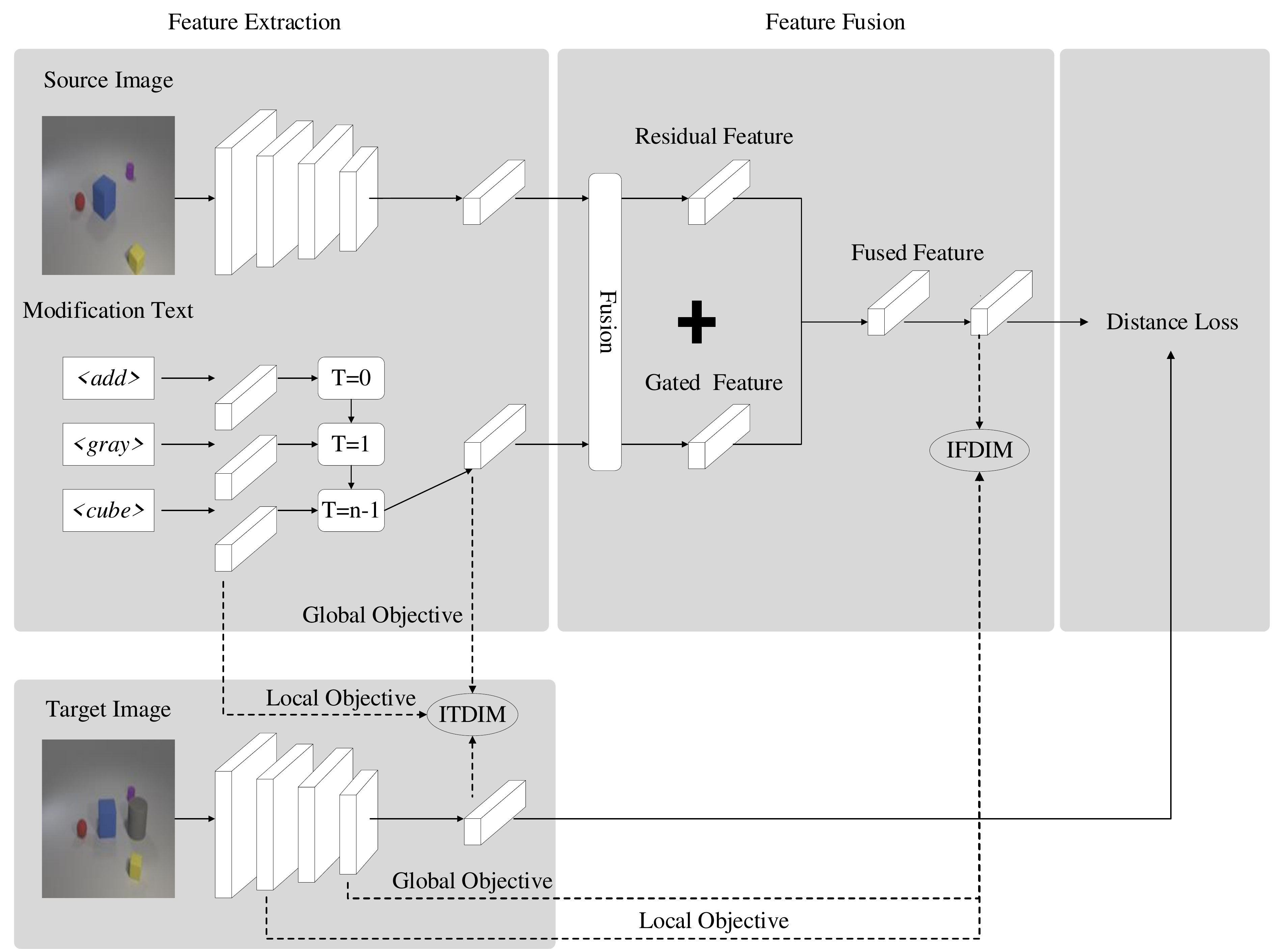}
\caption{The system pipeline for training based on the abstract modification text.}
\label{fig2}
\end{figure}
\section{Text Image Residual Gating Based on Deep Information Maximization}
Since the modality gap caused by the distributional difference of features between modalities significantly influences the cross-modal image retrieval accuracy, we erase this gap by applying mutual information maximization between the representations of the image, the text and their fusion. Figure \ref{fig2} is the system pipeline of our model.

\subsection{Feature Fusion Based on Deep Mutual Information Maximization}

In our task, there are two main modality gaps influencing our retrieval accuracy: 1)the gap between the source image and text which makes feature fusion insufficient; 2) the gap between the fusion and the desired image which directly affects the similarity learning. We utilize deep mutual information to narrow these two gaps in this and the next section.

The main task of the feature fusion module is to compose the semantic information extracted from the source image and the modified text. Since the fusion network's inputs are the features of the source image and the modified text, we first encode each input by a corresponding classical encoder.  

For the input images, we adopt a ResNet-18 whose output dimension of the last fc is changed to 512 to extract their features. For the modified text, we firstly embed it into a distributed embedding space and get the word embeddings; Then, we employ a widely used sequence learning model: Long Short-Term Memory (LSTM) to learn the sentence representations. We define the text's feature as the hidden state at the final time step.

After obtaining these features, we fuse them by the basic network of our method, Text Image Residual Gating (TIRG) \cite{vo2019composing}. TIRG is composed of gating feature and residual feature and it fuses image and text features by the following approach,
\begin{align}
\phi_{st}^{rg} = w_g f_{g}(\phi_s,\phi_t) + w_r f_{r}(\phi_s,\phi_t)
\label{eqution1}
\end{align}
\noindent where $f_g, f_r$ denote gating and the residual features presented in Figure \ref{fig2}. $w_g$ and $w_r$ are the trade-off between these two features. This gating feature in TIRG can be formulated as follows,
\begin{align}
f_{g}(\phi_s,\phi_t) = \sigma(\mathrm{FC_{g2}}(\mathrm{ReLU}(\mathrm{FC_{g1}}([\phi_s,\phi_t])))) \odot \phi_s
\label{eqution2}
\end{align}
\noindent where $\sigma$ is the sigmoid function, $\odot$ is element wise product, $\mathrm{FC}_{g1}$ and $\mathrm{FC}_{g2}$ represent fully connected layers and $[ {\phi_s,\phi_t} ]$ denotes the concatenation of $\phi_s$ and $\phi_t$. This feature is utilized to judge whether the modified text is helpful to the query image or not and retain the image feature if these two inputs are sufficiently different.
The residual feature is computed as follows,
\begin{align}
f_{r}(\phi_s,\phi_t) =\mathrm{FC_{r2}}( \mathrm{ReLU}(\mathrm{FC_{r1}}([\phi_s,\phi_t])))
\label{eqution3}
\end{align}
\noindent where $FC_{r1}$ and $FC_{r2}$ are fully connected layers.

Figure \ref{fig2} presents the model for the dataset with abstract modification text, such as Fashion200k and MIT-states, and we call it TIRG$_A$. For the dataset whose modification text is more concrete like CSS, we use the TIRG$_C$ model. TIRG$_C$ changes the feature extraction module to alter the spatial properties of the output of the image encoder. It replaces the source image encoder with ResNet-17 and broadcasts the text's feature along the height and width dimension to make it match the source image's feature. Accordingly, fully connected layers in the fusion network are replaced by convolutional layers with $3 \times 3$ filter kernels.

In our framework, distance metric learning optimizes the whole network by measuring the similarity between the desired image and the fusion of the source image and the modified text. As a classical encoder, ResNet-18 can get the desired image's high-quality feature in the fully labeled dataset \cite{he2016deep,shwartz2017opening}. Thus, the quality of the fusion feature is crucial to the retrieval accuracy. However, as the source image and the modified text are from different modalities, their features usually have inconsistent distribution and representation, which leads to a modality gap. The fusion network can not compose the semantic information captured from the source image and the modified text sufficiently without erasing the modality gap before fusing. Inspired by the recent advance of Deep InfoMax, we use mutual information maximization the image modality and the text modality (ITDIM) to narrow their modality gap. Considering that the source image and the modified text are completely semantically different, it's hard to narrow their modality gap by capturing the non-linear statistical dependencies using MI maximization between their representation \cite{belghazi2018mine}. Thus, ITDIM maximizes mutual information between the representation of the desired image and that of the modified text which contain partially same semantic information. In the previous work of the image to text retrieval, researchers attempt to obtain a better alignment of distributions of item representations across modalities by adversarial learning. They treat the input query encoder as the "generator" in GAN \cite{goodfellow2014generative} and design a modality classifier, which acts as the "discriminator". Given an unknown feature projection, this classifier detects the modality of an item as reliably as possible  \cite{wang2017adversarial,wang2019learning}. In essence, these methods are equivalent to maximizing the mutual information between the feature vectors of the different modalities with the same semantic information. Further, narrowing modality gap by adversarial loss can be viewed as the special cases of MINE \cite{belghazi2018mine}, the basic version of Deep Infomax. However, it is often unsufficient to quantify the dependency between two modalities by estimating mutual information between the two complete representations (i.e., feature vectors), namely global MI maximization. Rather, combining the average MI maximization between the high-level representation and local regions of the low-level representation (e.g., patches rather than the complete representation) \cite{hjelm2018learning}, namely local MI maximization, can get a better distribution alignment. When the representation is the outputs of different layers of the same encoder, the local MI maximization makes the encoder prefer information that is shared among patches and filter noise specific to local patches. For our task, each modality makes use of different encoders and the semantic information the modified text contains is part of that the desired image contains. If we maximize the local MI between the high-level representation of the modified text and the low-level representation of the desired image, the desired image encoder will discard some image-specific semantic information as noise. Thus, our ITDIM maximizes mutual information between the high-level representation in the desired image encoder and the low-level representation in the modified text encoder. We verify the said analysis through experiments in section 4.5.3.

In this paper, we maximize MI using different MI($X; Y$) objectives, where $X$ is a low-level feature map, and $Y$ is a high-level feature vector \cite{hjelm2018learning}. Generally speaking, mutual information quantifies the dependence of $X$ and $Y$. we formulate it as follows,
\begin{equation}
    I(X;Y)=\int_{\mathcal{X} \times \mathcal{Y}} \log \frac{d \mathbb{P}_{XY}}{d \mathbb{P}_{X} \otimes \mathbb{P}_{Y}} d \mathbb{P}_{XY} = \int_{\mathcal{X} \times \mathcal{Y}} \log \frac{d \mathbb{P}_{XY/X}}{d \mathbb{P}_{Y}} d \mathbb{P}_{XY}
    \label{eqution4}
\end{equation}
where $ \mathbb{P}_{XY}$ is the joint probability distribution, and $\mathbb{P}_X= \int_{\mathcal{Y}} \mathbb{P}_{XY}$ and $\mathbb{P}_Y= \int_{\mathcal{X}} \mathbb{P}_{XY}$ are the marginals \cite{belghazi2018mine}. In the original Deep InfoMax, $X$ and $Y$ are the different-level representations in an encoder and contain the same semantic information. According the Equation (\ref{eqution4}), the mutual information maximization makes $Y$ capture the representative information of $X$ as much as possible. When we set $X$ and $Y$ as the representations of the modified text and the desired image, ITDIM actually guarantees the image representation to hold the semantic information related to the modified text as much as possible.

To make the following section more clear, we give some notations. We define the modified text, the source image, the desired image and their features as$t$, $s$, $d$, $\phi_t$, $\phi_s$ and $\phi_d$, respectively. The fusion feature is denoted by $\phi_{st}$.  Then we define the input image as $I \in (s,d)$ and its feature as $\phi_I \in (\phi_s,\phi_d)$. In the image encoder, the representation is defined as $i = I_{m}(I,\theta_{im})$ where $I_{m}$ denotes the image CNN before $i$ and $\theta_{im}$ denotes the parameters of this CNN. To obtain the best retrieval accuracy, we set $i$ to varied layers in terms of different MI objectives and datasets. In the text encoder, we set each representation as $e = E_m(t,\theta_{tm})$, which $E_m$ is the LSTM network with parameters $\theta_{tm}$. The text encoder in our method consists of two representations: the output and the parallel connection of the word embeddings.

The key to maximize the MI is to design an appropriate MI estimator. Noise-Contrastive Estimation \cite{gutmann2010noise,gutmann2012noise} and Donsker-Varadhan estimators \cite{donsker1983asymptotic} require a large number of negative samples to be competitive and quickly becomes cumbersome with increasing batch size. By contrast, Jensen-Shannon MI estimator \cite{hjelm2018learning,nowozin2016f} performs well using a small quantity of negative samples,  so we apply this estimator to our model. Our estimator $\widehat{\mathcal{I}}_{\theta_d,\theta_{tm},\theta_{im}}^{\mathrm{JSD}}$ for ($e;i$), $i = \phi_d$ can be formulated as,
\begin{align}
 \widehat{\mathcal{I}}_{\theta_d,\theta_{tm},\theta_{im}}^{\mathrm{JSD}}(e;i) & := \mathbb{E}_{\mathbb{P}_e}[\mathrm{-sp}(-T_{\theta_d,\theta_{tm},\theta_{im}}(e,i))] - {\mathbb{E}_{\mathbb{P}_e \times \mathbb{P}_{e^{'}}}} [\mathrm{sp} (T_{\theta_d,\theta_{tm},\theta_{im}}(e^{'},i))]
\label{eqution5}
\end{align}
\noindent where $\mathbb{P}_e$ is the empirical probability distribution of text representation $e$, $e^{'}$ is the low-level representation sampled from $\mathbb{P}_{e^{'}}$ = $\mathbb{P}_e$, $\mathrm{sp}(z) = \log(1+e^z)$ and $T$ can be concretized as a discriminator function modeled by deep neural network with parameters $\theta_d$. 

To compute the mutual information between high dimensional representation pairs effectively and sufficiently, we maximize MI by adopting global MI objectives and local MI objective, which maximizes the MI between the complete $X$ and $Y$ and estimates the MI between $Y$ and local regions of $X$ and respectively. Based on our estimator, we define our global MI and local MI objectives as follows,
\begin{align}
    MI_E^G(e;i) = \max \limits_{\theta_{dg},\theta_{tm},\theta_{im}} \widehat{\mathcal{I}}_{\theta_{dg},\theta_{tm},\theta_{im}}^{\mathrm{JSD}} (e;i)
    \label{eqution6}
\end{align}
\begin{align}
    MI_E^L(e;i) = \max \limits_{\theta_{dl},\theta_{tm},\theta_{im}} \frac{1}{M^2} \sum_{i=1}^{M^2} \widehat{\mathcal{I}}_{\theta_{dl},\theta_{tm},\theta_{im}}^{\mathrm{JSD}} (e^p;i)
    \label{eqution7}
\end{align}
\noindent where $\theta_{dg}$ and $\theta_{dl}$ are the parameters of the discriminators for the global and local MI objectives and $e^p$ is the $p$th patch of the feature map $e$.

Besides increasing the dependence between the text modality and the image modality, we also improve the compactness of the image feature by imposing prior matching objective, which makes $Y$ match a prior distribution. This objective can be formulated as,
\begin{align}
MI_E^P(i) = \mathbb{E}_{\mathbb{V}_y}[\log\mathcal{D}_{\theta_{dp}}(y)] + \mathbb{E}_{\mathbb{P}_e}[\log(1-\mathcal{D}_{\theta_{dp}}(i)]
\label{eqution8}
\end{align}
\noindent where $y$ denotes a random variable with prior probability distribution $\mathbb{V}_y$ and ${\theta_{dp}}$ is the parameters of the discriminator function $\mathcal{D}_{\theta_{dp}}$ used in this objective. Finally, we utilize these three objectives together and get the complete objective, 
\begin{align}
L_{E} = MI_E(e;i) = \alpha MI_E^G(e;i) + \beta MI_E^L(e;i) + \gamma MI_E^P(i)
\label{eqution9}
\end{align}
\noindent where $\alpha$, $\beta$ and $\gamma$ are the trade-off parameters.

The discriminators in MI objectives vary according to different application scenarios \cite{hjelm2018learning}. In our method, we define the discriminators for the global, local and prior matching objectives as Table \ref{table1}. We set the unit number of most hidden layers to the dimension of each encoder's output, $512$. Since the semantic information in the modified text is a portion of that in the desired image, there may be different text corresponding to the same image. And it's hard to determine which text-image pairs should have more mutual information. Thus, the 'fake' sample $e^{'}$ in Equation (\ref{eqution5}) is set as the same low-level feature as the 'real' sample $k$ extracted from another text that is not the description of the desired image.

\begin{table}[ht]
    \centering
    \begin{tabular}{|c|c|c|c|}
    \hline
    Objective  & Operation  & R@size  & Activation\\
    \hline
    \multirow{3}{*}{Global}
    & Input $\rightarrow$ Linear layer       & 512  & ReLU\\
    & Linear layer                           & 512  & ReLU\\
    & Linear layer                           & 1    &\\
    \hline
    \multirow{3}{*}{Local}
    & Input $\rightarrow$ 1 $\times$ 1 conv  & 512  & ReLU\\
    & 1 $\times$ 1 conv                      & 512  & ReLU\\
    & 1 $\times$ 1                           & 1    &\\
    \hline
    \multirow{3}{*}{Prior}
    & Input $\rightarrow$ 1 $\times$ 1 conv  & 512  & ReLU\\
    & 1 $\times$ 1 conv                      & 300  & ReLU\\
    & 1 $\times$ 1                           & 1    &\\
    \hline
    \end{tabular}
    \caption{Network architecture for global DIM,local DIM and prior matching}
    \label{table1}
    \end{table} 

\subsection{Distance Metric Learning based on Deep Mutual Information Maximization}

The goal of deep metric learning (DML) is to push closer the fusion feature $\phi_{st}$ and the desired image's feature $\phi_d$ while pulling apart the non-similar image's feature $\phi_n$. More precisely, suppose the training minibatch has $B$ queries, we select one fusion feature $\phi_{st}^t$ and create a corresponding set $\mathcal{N}_i$ that consists of a desired image $\phi_d^t$ and $K-1$ non-similar images $\phi_n^1,...,\phi_n^{K-1}$. We repeat this selection $M$ times and denote the $m$th selction as $\mathcal{N}_i^m$. We adopt the following softmax cross-entropy loss,
\begin{align}
L_{T} = -\frac{1}{MB}\sum_{t=1}^B \sum_{m=1}^M \log\left\{\frac{\exp\{\kappa(\phi_{st}^t,\phi_d^t)\}}{\sum_{\phi_a \in \mathcal{N}_i^m}\exp\{\kappa(\phi_{st}^t,\phi_a)\}} \right\}
\label{eqution10}
\end{align}
\noindent where $\kappa$ is a similarity kernel and can be implemented as the dot product or negative $l_2$ distance. If we apply large $K$ to the above equation, each desired image is contrasted with a lot of other non-similar images. Our model becomes more discriminative and fits faster than that using small $K$, but can be more apt to overfitting. Hence, we adopt this model to the dataset which is difficult to converge. In our experience, we use $K=B$ and $M=1$ for Fashion200k and the Equation (\ref{eqution11}) can be rewritten as follows,
\begin{align}
L_{T} = -\frac{1}{B} \sum_{t=1}^B \log\left\{\frac{\exp\{\kappa(\phi_{st}^t,\phi_d^t)\}}{\sum_{j=1}^B \exp\{\kappa(\phi_{st}^t,\phi_d^j)\}} \right\}.
\label{eqution11}
\end{align}
By contrast, $K$ can also be set very small. In the extreme case, when we use the smallest value of $K=2$, the loss is the same as the soft triplet loss in the previous literature \cite{vo2016localizing,hermans2017defense}. The loss function can be formulated as follows, 
\begin{align}
L_{T} = \frac{1}{MB} \sum_{t=1}^B \sum_{m=1}^M \log \{1+\exp\{\kappa(\phi_{st}^m,\phi_n^{m,t})-\kappa(\phi_{st}^m,\phi_d^t)\}\}
\label{eqution12}
\end{align}
where $\phi_n^{m,t}$ denotes the $t$th fusion feature in the $\mathcal{N}_i^m$. This loss function is applied to the other two datasets, namely MIT-States and CSS.

Compared with the DML in the unimodal scenario \cite{weinberger2009distance,gu2019local,oh2016deep,wang2019multi}, the precondition of the similarity learning between different modalities is to learn a common subspace where the items of different modalities can be directly compared to each other. As the inputs of the DML are the fusion feature and the desired image feature, there is also a modality gap existing due to their inconsistent distribution. Compared to the modality gap between the image and the text in the last section, the modality gap between the fusion and the image is smaller because most of the semantic information in the fusion modality comes from the image modality. Hence, the distributions of the features of the image and the fusion are similar to some extent. If we want to get better retrieval performance, we need to improve the similarity until these two distributions are highly consistent. We achieve this goal by maximizing the mutual information between the representations of the fusion and the desired image, which contains the same semantic information. Since TIRG obtains the fusion feature by adding the gating feature and the residual feature, no feature map which contains all semantic information in the fusion network can be used as low-level representation in the local MI objective. If we maximize MI between the high-level layer in the gating or the residual network which contains partial semantic information of the desired image and the low-level layer in the desired image encoder, local MI objective will discard partial semantic information that is unique to desired image as noise. Thus, we maximize mutual information between the low-level representation in the desired image encoder and the high-level representation in the fusion network. Furthermore, experiments in section 4.5.3 demonstrate that using different layers in the desired image encoder as $X$ in the global and local MI objectives is much better than using fusion feature ($\phi_{st})$) as $X$ in global MI objective. Thus, we set $Y$ in MI($X; Y$) as the high-level layer in the fusion network and $X$ as the low-level layer in the desired image's encoder, $i = I_{m}(I,\theta_{im})$, which is defined in the last section. The mutual information maximization here is between the image modality and the fusion modality, so we call it IFDIM. 

The setting of $X$ and $Y$ makes IFDIM optimizes the parameters of the whole architecture. We define $\theta_a$ as the parameters of the entire model and our cross-modal Jensen-Shannon MI estimator can be written as,
\begin{align}
\widehat{\mathcal{I}}_{\theta_d,\theta_{a}}^{\mathrm{CJ}}(i;\phi_{st}) := \mathbb{E}_{\mathbb{P}_i}[\mathrm{-sp}(-T_{\theta_d,\theta_{a}}(i,\phi_{st}))] - {\mathbb{E}_{\mathbb{P}_i \times \mathbb{P}_{i^{'}}}} [\mathrm{sp} (T_{\theta_d,\theta_{a}}(i^{'},\phi_{st}))]
\label{eqution13}
\end{align}
where $\mathbb{P}_i$ is the empirical probability distribution of $i$, $\mathbb{P}_{i'} = \mathbb{P}_i$ is the distribution of $i'$ and $T$ stands for a discriminator function with parameters $\theta_d$ as Equation (\ref{eqution5}). As section 3.1, we increase the dependence between the fusion modality and the image modality by global MI, local MI and prior matching objectives. Since the formulas of these objectives are similar and can be obtained by altering corresponding parameters, we directly provide the complete objective,
\begin{align}
MI_F(i;\phi_{st}) = \alpha MI_F^G(i;\phi_{st}) + \beta MI_F^L(i;\phi_{st}) + \gamma MI_F^P(\phi_{st})
\label{eqution14}
\end{align}
\noindent where $\alpha$, $\beta$ and $\gamma$ are trade-off parameters defined in Equation (\ref{eqution5}). And the loss function for the cross-modal Deep InfoMax can be represented as $L_F = MI_F(i;\phi_{st})$.
 
Finally, we train our model by the overall loss function defined as,
\begin{align}
L_{ALL} =\mu (L_{E} + L_{F}) + L_{T}
\label{eqution15}
\end{align}
\noindent where $\mu$ is dynamic tradeoff hyperparameters. 

\section{Experiments}
This section consists of three parts: 1) introduce the experimental settings; 2) compare our method with the state-of-the-art algorithms on different datasets; 3) provide ablation experiments to study the effect of the ITDIM and the IFDIM in our model.

\subsection{Experimental Settings}
We compare our method with TIRG \cite{vo2019composing} and seven benchmarks mentioned in section 2.1 on three datasets: Fashion200k \cite{han2017automatic}, MIT-States \cite{isola2015discovering} and CSS \cite{vo2019composing}. Our main metric for retrieval is recall at rank k (R@k), computed as the percentage of the text queries where (at least 1) desired or correct labeled image is within the top K retrieval images. In order to get stable retrieval results, we repeat each experiment $5$ times, and both mean and standard deviation are reported. We use PyTorch in our experiments. For all datasets, the low-level representation in $MI(X; Y)$ objectives used in the image encoder are set as the last convolutional layer for its better performance. By default, training is run for 160k iterations with a start learning rate 0.01. We will release the code to the public. The weights $\alpha$, $\beta$ and $\gamma$ are set as 0.5, 1 and 0.1. We set $\mu = {L_T}/{15(L_E+L_F)}$ with initial value 0.001 and update it every 10k iterations. We apply our method TIRG-DIM$_A$ to Fashion200k and MIT-States and TIRG-DIM$_C$ to CSS in terms of their modified text's attribute.   
\subsection{Fashion200k}

\begin{table}[t]
\centering
\begin{tabular}{llll}
\toprule
Method&R@1&R@10&R@50\\
\hline
\cite{han2017automatic}  &6.3  &19.9  &38.3\\
Image only  & 3.5  & 22.7  & 43.7\\
Text only   & 1.0  & 12.3  & 21.8\\
Concatenation  & $11.9^{\pm 1.0}$  & $39.7^{\pm 1.0}$  & $62.6^{\pm 0.7}$\\
Show and Tell  & $12.3^{\pm 1.1}$  & $40.2^{\pm 1.7}$  & $61.8^{\pm 0.9}$\\
Param Hashing  & $12.2^{\pm 1.1}$  & $40.0^{\pm 1.1}$  & $61.7^{\pm 0.8}$\\
Relationship   & $13.0^{\pm 0.6}$  & $40.5^{\pm 0.7}$  & $62.4^{\pm 0.6}$\\
Film           & $12.9^{\pm 0.7}$  & $39.5^{\pm 2.1}$  & $61.9^{\pm 1.9}$\\
TIRG           & $\underline{14.1}^{\pm 0.6}$ & $\underline{42.5}^{\pm 0.7}$ & $\underline{63.8}^{\pm 0.8}$\\
\hline
TIRG-DIM$_A$     & $\mathbf{17.4}^{\pm 0.3}$ & $\mathbf{43.4}^{\pm 0.4}$ & $\mathbf{64.5}^{\pm 0.6}$\\
\bottomrule
\end{tabular}
\caption{Retrieval performance on Fashion200k. The best result is in bold and the second best in underline.}
\label{table2}
\end{table} 

\begin{figure}[htbp]
    \centering
    \includegraphics[scale=0.45]{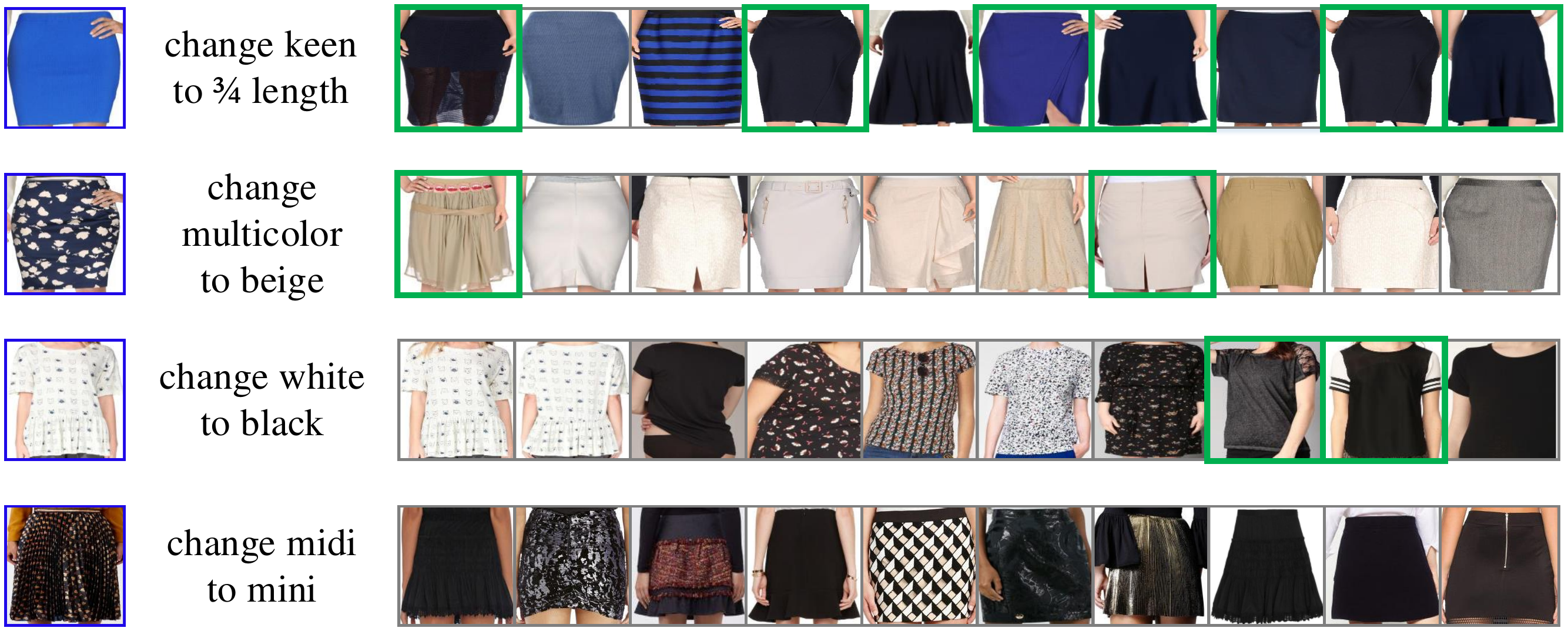}
    \caption{Qualitative results of image retrieval with modified text on Fashion200k. blue/green boxes: source/desired images.}
    \label{fig3}
\end{figure}

\begin{figure}[htbp]
    \centering
    \includegraphics[scale=0.45]{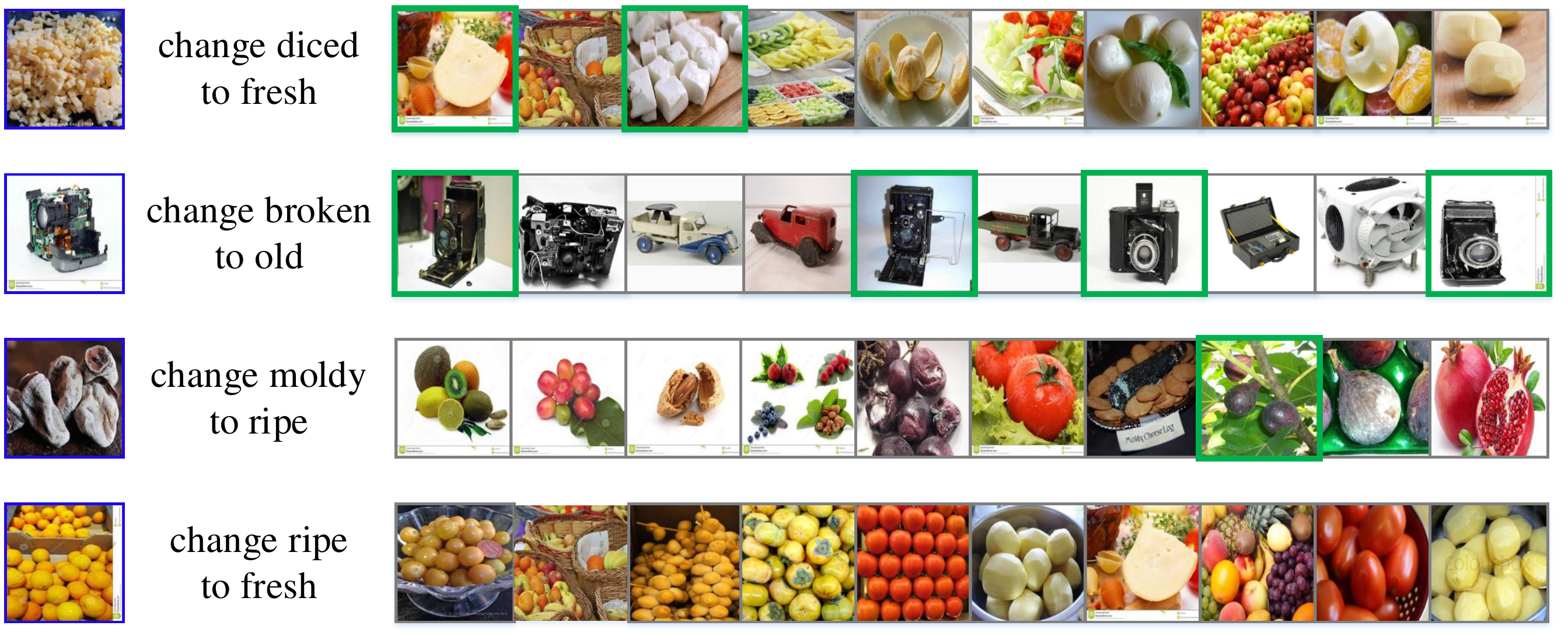}
    \caption{Qualitative results of image retrieval with modified text on MIT-states. blue/green boxes: source/desired images.}
    \label{fig4}
\end{figure}

Fashion200k is a widely-used dataset in the filed of cross-modal image retrieval. It is composed of 200k images of fashion products and each image has a compact attribute-like description (such as mini and short dress or knee length skirt). Following the previous work \cite{han2017automatic}, queries are generated as follows: the query images and its desired images have one word difference in their descriptions, and the modified text is this different word. We adopt the same training split as TIRG \cite{vo2019composing} and generate queries on the fly for training. We randomly sample $10$ validation sets of $3167$ test queries and report the mean. 

Figure \ref{fig3} illustrates some qualitative results and Table \ref{table2} shows the retrieval accuracy on this dataset. From the results we have the following observations: 1) our method outperforms all the other approaches with a large margin, especially the R@1 performance, which has a more than 23 percent increase over the best competitor; 2) The standard deviations of the proposed method are smaller than others. These observations demonstrate that we can get more accurate and stable retrieval performance by improving the quality of each input's feature and the fusion feature using mutual information maximization.

\begin{table}[t]
\centering
\begin{tabular}{llll}
\toprule
Method  & R@1  & R@5  & R@10\\
\hline
Image only       & 3.3  & 12.8  & 20.9\\
Text only        & 7.4  & 21.5  & 32.7\\
Concatenation    & $11.8^{\pm 0.2}$ & $30.8^{\pm 0.2}$  & $42.1^{\pm 0.3}$\\
Show and Tell    & $11.9^{\pm 0.1}$ & $31.0^{\pm 0.5}$  & $42.0^{\pm 0.8}$\\
Att. as Operator & $8.8^{\pm 0.1}$  & $27.3^{\pm 0.3}$  & $39.1^{\pm 0.3}$\\
Relationship     & $\underline{12.3}^{\pm 0.5}$  & $\underline{31.9}^{\pm 0.7}$  & $42.9^{\pm 0.9}$\\
Film             & $10.1^{\pm 0.3}$ & $27.7^{\pm 0.7}$  & $38.3^{\pm 0.7}$\\
TIRG             & $12.2^{\pm 0.4}$ & $\underline{31.9}^{\pm 0.3}$  & $\underline{43.1}^{\pm 0.3}$\\
\hline
TIRG-DIM$_A$       & $\mathbf{14.1}^{\pm 0.3}$  & $\mathbf{33.8}^{\pm 0.5}$  & $\mathbf{45.0}^{\pm 0.5}$ \\
\bottomrule
\end{tabular}
\caption{Retrieval performance on MIT-States. The best result is in bold and the second best in underline.}
\label{table3}
\end{table}

\subsection{MIT-States}

MIT-States has $63440$ images, and each image is described by an object/noun word and a state/adjective word (such as wide belt or tiny island). In total, this dataset contains $245$ nouns and $115$ adjectives and each individual noun is only modified by ~9 adjectives it affords.

For image retrieval, we create the query image and desired image by sampling pairs of images with the same object labels and different state labels. The state of the desired image is considered as the modified text. Therefore, our method is to retrieve image which possesses the same object but new state compared with the query image. In the experiments, we select 80 nouns for training, and the others are adopted for testing. Based on these settings, models are trained by different state/adjective (modified text) and tested using unseen objects.

A number of qualitative results are shown in Figure \ref{fig4} and the quantitative retrieval results can be seen from Table \ref{table3}. Obviously, our proposed method obtains the highest retrieval accuracy at different R@k on this dataset. More specifically, we achieve $15\%$ and $6\%$ improvement on R@1 and R@10 respectively compared with the second best algorithm, namely Relationship \cite{santoro2017simple}. Because the same object with varied states can look extremely different, the modification text becomes more significant. Therefore, the ``Text only" baseline overcomes ``Image only".
\subsection{CSS dataset}
\begin{figure}[htbp]
    \centering
    \includegraphics[scale=0.45]{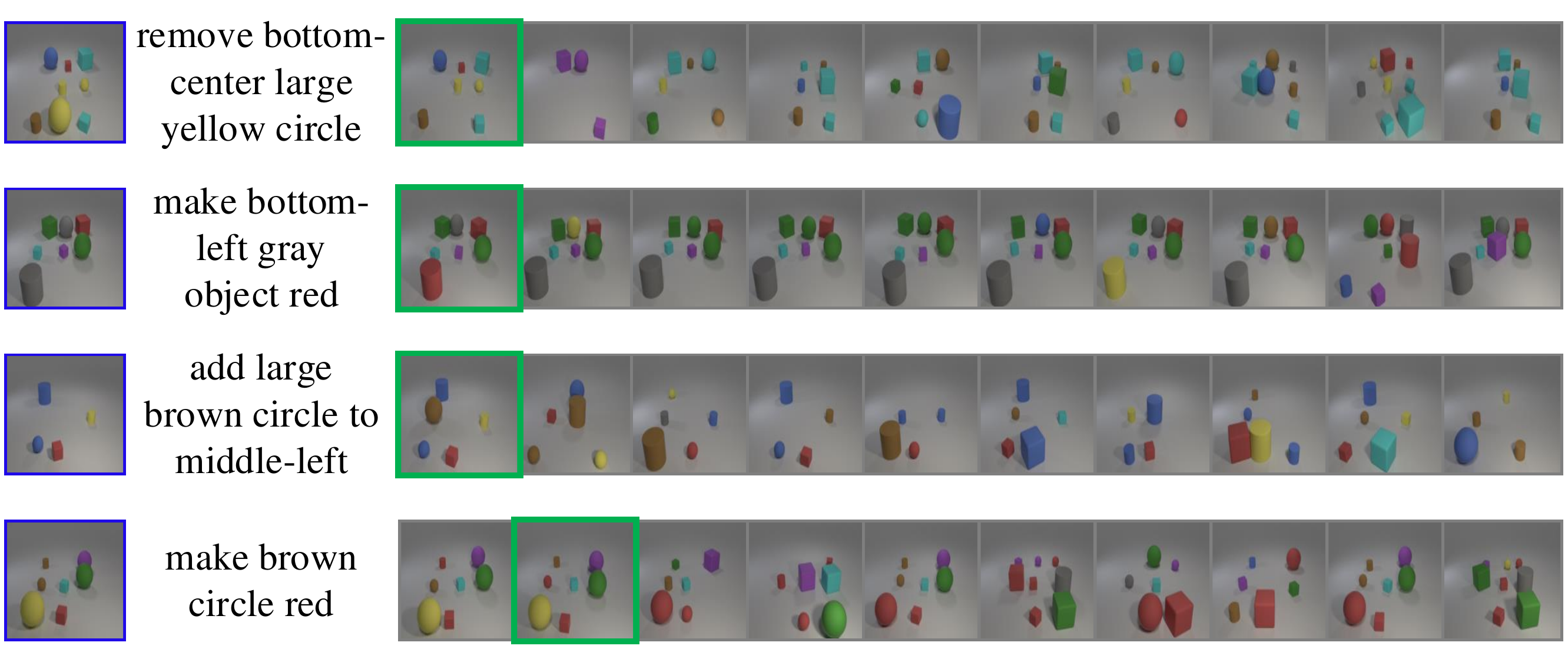}
    \caption{Qualitative results of image retrieval with modified text on CSS. blue/green boxes: source/desired images.}
    \label{fig5}
\end{figure}

\begin{table}[t]
\centering
\begin{tabular}{llll}
\toprule
Method  & R@1  & R@5  & R@10\\
\hline
Image only       & $6.3$  & $29.3$  & $54.0$\\
Text only        & $0.1$  & $0.5$  & $0.8$\\
Concatenation    & $60.6^{\pm 0.5}$ & $88.2^{\pm 0.4}$  & $92.8^{\pm 0.4}$\\
Show and Tell    & $33.0^{\pm 3.2}$ & $75.0^{\pm 1.3}$  & $83.0^{\pm 0.9}$\\
Para.Hasing      & $60.5^{\pm 1.9}$  & $88.1^{\pm 0.8}$  & $92.9^{\pm 0.6}$\\
Relationship     & $62.1^{\pm 1.2}$  & $89.1^{\pm 0.4}$  & $93.5^{\pm 0.7}$\\
Film             & $65.6^{\pm 0.5}$ & $89.7^{\pm 0.6}$  & $94.1^{\pm 0.5}$\\
TIRG             & $\underline{73.7}^{\pm 0.4}$ & $\underline{90.7}^{\pm 0.4}$  & $\underline{94.6}^{\pm 0.4}$\\
\hline
TIRG-DIM$_A$       & $\mathbf{77.0}^{\pm 0.2}$  & $\mathbf{95.6}^{\pm 0.4}$  & $\mathbf{97.6}^{\pm 0.3}$ \\
\bottomrule
\end{tabular}
\caption{Retrieval performance on CSS. The best result is in bold and the second best in underline.}
\label{table4}
\end{table}

CSS consists of 32k synthesized images in a 3-by-3 grid scene which are generated by CLEVR toolkit. Objects in the images are rendered with different color, shape and size occupy. Each image comes in a simple 2D blobs version and 3D version, and we utilize the second one in this paper. There are 16k queries for training and 16k queries for testing in this dataset. Each query is composed of a source image, a modified text and a desired image (Figure \ref{fig1}). The modification text has three templates: adding, removing or changing object attributes, such as "add small green rectangle to top-right", "remove bottom-center small red circle" or "make bottom-left large green object gray". We provide a stronger test of generation by making certain object shape and color combinations only appear in training and not in testing, and vice versa.

We can find the quantitative and qualitative results from Figure \ref{fig5} and Table \ref{table4} respectively. All the methods except the Image Only and the Text Only approaches get much higher retrieval accuracy on this dataset than on the other two. We believe this is because the image queries are simple and the text queries contain more information. Compared to the second best method TIRG, TIRG-DIM$_{A}$ improves retrieval accuracy by 3.3, 4.9 and 3.0 percentage points on R@1, R@5 and R@10 score.

\subsection{Ablation Studies}
In this section, we first provide the R@1 accuracy of various ablation studies to gain insight into which part of our method matters the most. The results are in Table \ref{table5}. Then we offer the loss values and the visualization of distribution by Figure \ref{fig6} and Figure \ref{fig7}.

\begin{table}[ht]
\centering
\begin{tabular}{llll}
\toprule
Method  & Fashion  & MIT-State  & CSS\\
\hline
TIRG$_A$               & $14.1^{\pm 0.6}$  & $12.2^{\pm 0.4}$  & $71.2^{\pm 0.4}$\\
TIRG$_A$ + DIM$_{TextSour}$  & $13.6^{\pm 0.6}$  & $11.6^{\pm 0.5}$  & $70.5^{\pm 0.6}$\\
TIRG$_A$ + DIM$_{SourText}$  & $13.7^{\pm 0.4}$  & $11.7^{\pm 0.6}$  & $70.7^{\pm 0.5}$\\
TIRG$_A$ + DIM$_{SourDes}$  & $13.5^{\pm 0.5}$  & $11.4^{\pm 0.4}$  & $70.4^{\pm 0.5}$\\
TIRG$_A$ + DIM$_{DesSour}$  & $13.3^{\pm 0.7}$  & $11.3^{\pm 0.5}$  & $70.2^{\pm 0.4}$\\
TIRG$_A$ + DIM$_{DesText}$  & $13.1^{\pm 0.6}$  & $11.2^{\pm 0.4}$  & $70.1^{\pm 0.5}$\\
TIRG$_A$ + DIM$_{TextFus}$  & $14.2^{\pm 0.6}$  & $12.3^{\pm 0.4}$  & $71.4^{\pm 0.3}$\\
TIRG$_A$ + DIM$_{SourFus}$  & $14.3^{\pm 0.5}$  & $12.4^{\pm 0.3}$  & $71.3^{\pm 0.2}$\\
TIRG$_A$ +DIM$_{FusDes}$  & $14.5^{\pm 0.4}$ & $12.5^{\pm 0.3}$  & $71.6^{\pm 0.3}$\\
TIRG$_A$ +DIM$_{ResiDes}$ & $14.7^{\pm 0.5}$  & $12.6^{\pm 0.4}$  & $71.7^{\pm 0.3}$\\
TIRG$_A$ +DIM$_{GatingDes}$  & $14.8^{\pm 0.4}$  & $12.6^{\pm 0.3}$  & $71.8^{\pm 0.3}$\\
TIRG$_A$ + ITDIM        & $15.4^{\pm 0.4}$  & $12.7^{\pm 0.3}$  & $72.1^{\pm 0.2}$\\
TIRG$_A$ + IFDIM        & $16.5^{\pm 0.3}$  & $13.7^{\pm 0.2}$  & $73.2^{\pm 0.3}$\\
TIRG-DIM$_A$           & $17.4^{\pm 0.3}$  & $14.1^{\pm 0.3}$  & $73.8^{\pm 0.2}$\\
\hline
TIRG$_C$              & $12.4^{\pm 0.5}$  & $10.3^{\pm 0.5}$  & $73.7^{\pm 0.4}$\\
TIRG$_C$ + DIM$_{TextSour}$ & $11.8^{\pm 0.6}$  & $9.9^{\pm 0.5}$  & $73.1^{\pm 0.5}$\\
TIRG$_C$ + DIM$_{SourText}$ & $12.0^{\pm 0.5}$  & $10.0^{\pm 0.4}$  & $73.3^{\pm 0.6}$\\
TIRG$_C$  + DIM$_{SourDes}$ & $11.6^{\pm 0.5}$  & $9.8^{\pm 0.5}$  & $73.1^{\pm 0.5}$\\
TIRG$_C$  + DIM$_{DesSour}$ & $11.5^{\pm 0.6}$  & $9.6^{\pm 0.4}$  & $72.9^{\pm 0.5}$\\
TIRG$_C$  + DIM$_{DesText}$ & $11.4^{\pm 0.5}$  & $9.5^{\pm 0.4}$  & $72.7^{\pm 0.5}$\\
TIRG$_C$ + DIM$_{TextFus}$  & $12.6^{\pm 0.4}$  & $10.4^{\pm 0.4}$  & $74.0^{\pm 0.4}$\\
TIRG$_C$ + DIM$_{SourFus}$ & $12.6^{\pm 0.3}$  & $10.5^{\pm 0.5}$  & $73.9^{\pm 0.3}$\\
TIRG$_C$ +DIM$_{FusDes}$  & $12.7^{\pm 0.4}$ & $10.7^{\pm 0.5}$  & $74.1^{\pm 0.3}$\\
TIRG$_C$ +DIM$_{ResiDes}$ & $12.9^{\pm 0.3}$  & $10.8^{\pm 0.4}$  & $74.3^{\pm 0.3}$\\
TIRG$_C$ +DIM$_{GatingDes}$  & $13.0^{\pm 0.3}$  & $11.0^{\pm 0.3}$  & $74.5^{\pm 0.3}$\\
TIRG$_A$ + ITDIM        & $15.4^{\pm 0.4}$  & $12.7^{\pm 0.3}$  & $72.1^{\pm 0.2}$\\
TIRG$_C$ + IFDIM        & $13.9^{\pm 0.3}$  & $12.3^{\pm 0.2}$  & $76.5^{\pm 0.3}$\\
TIRG-DIM$_C$           & $14.8^{\pm 0.2}$  & $12.9^{\pm 0.1}$  & $77.0^{\pm 0.2}$  \\
\hline
Our Full Model         & $\mathbf{17.4}^{\pm 0.3}$  & $\mathbf{14.1}^{\pm 0.3}$  & $\mathbf{77.0}^{\pm 0.2}$  \\
\bottomrule
\end{tabular}
\caption{Retrieval performance (R@1) of ablation studies}
\label{table5}
\end{table}

\subsubsection{Effect of ITDIM}

We study the effect of the deep mutual information maximization between low-level representation in the text encoder and the hihg-level representation in the desired image encoder (ITDIM) in both TIRG$_A$ and TIRG$_C$. From the results in Table \ref{table5}, we can see that the performance of the two models has a remarkable improvement by adding ITDIM into these models. It demonstrates that the ITDIM can help to get a better alignment of distributions of item representations between the modified text and the image, though the semantic information in the text modality is much less than that in the image modality. 

\subsubsection{Effect of IFDIM}

Comparing the results obtained with deep mutual information maximization between the low-level representation in the desired image encoder and the high-level representation in the fusion network (IFDIM) and that with ITDIM, we can see the models based on IFDIM leads to more considerable improvement. We believe this is because the representations of these two modalities all contain rich semantic information. The performance of DIM in our model is partly related to the quantity of the semantic information contained in its two inputs. 

\subsubsection{Effect of other Deep InfoMaxs}
In order to give a more detailed comparison between different deep InfoMaxs, we also provide experiments based on deep mutual information maximization using other low-level and high-level representation in Table \ref{table5}. 

To obtain better fusion features, we attempt to make us of DIM$_{TextSour}$, which maximize mutual information between the low-level representation in the text encoder and the high-level representation in the source image encoder, and DIM$_{SourText}$, which swaps the positions of the representation in MI objectives. As the low-level and high-level representation in these InfoMaxs are semantically different, it's harmful to retrieval to force Deep InfoMax to capture non-linear statistical dependencies between two semantically independent modalities. We also employ DIM$_{SourDes}$, DIM$_{DesSour}$ and DIM$_{DesText}$ (Des in the subscripts represents the representation in the the desired image encoder), whose low-level representation contain partially different semantic information compared to their high-level representation. In these InfoMaxs, local mutual information objectives discard part of semantic information that is unique to the low-level representation as mentioned in section 3.1, which affect the retrieval results. Moreover, we try to optimize the fusion network by aligning the distribution between the text representation and the fusion feature by DIM$_{TextFus}$ and narrowing the modality gap between the source image and the fusion feature by DIM$_{SourFus}$. The experimental results show that these two InfoMaxs marginally improve the performance. We think this is because the effect of mutual information maximization between the low-level representation and the high-level representation of the same deep neural network is limited in fully supervised learning. However, ITDIM can significantly improve the retrieval accuracy by estimating mutual information between the representation from different encoders. 

For distance metric learning, learning a common space between fusion feature and the desired image feature is the key to guarantee the items of different modalities can be directly compared to each other. Apart from IFDIM, we exploit Deep InfoMax between the low-level representation in the fusion network and the high-level representation in the desired image encoder. Considering that fusion feature in TIRG is the weighted sum of the residual feature and the gating feature, no feature map which contains all semantic information in the fusion network can be used as low-level representation in the local MI objective. Here we conduct experiments based on DIM$_{FusDes}$, DIM$_{ResiDes}$ and DIM$_{GatingDes}$, which give up using the local MI objective, apply feature map in the residual network and feature map in the gating work to local MI objective respectively. Although these Deep InfoMaxs can't compete with ITDIM and IFDIM, they overcome the others.  

\subsubsection{Qualitative Analyses of the Effects of ITDIM and IFDIM}

\begin{figure}[htbp]
    \centering
    \includegraphics[scale=0.7]{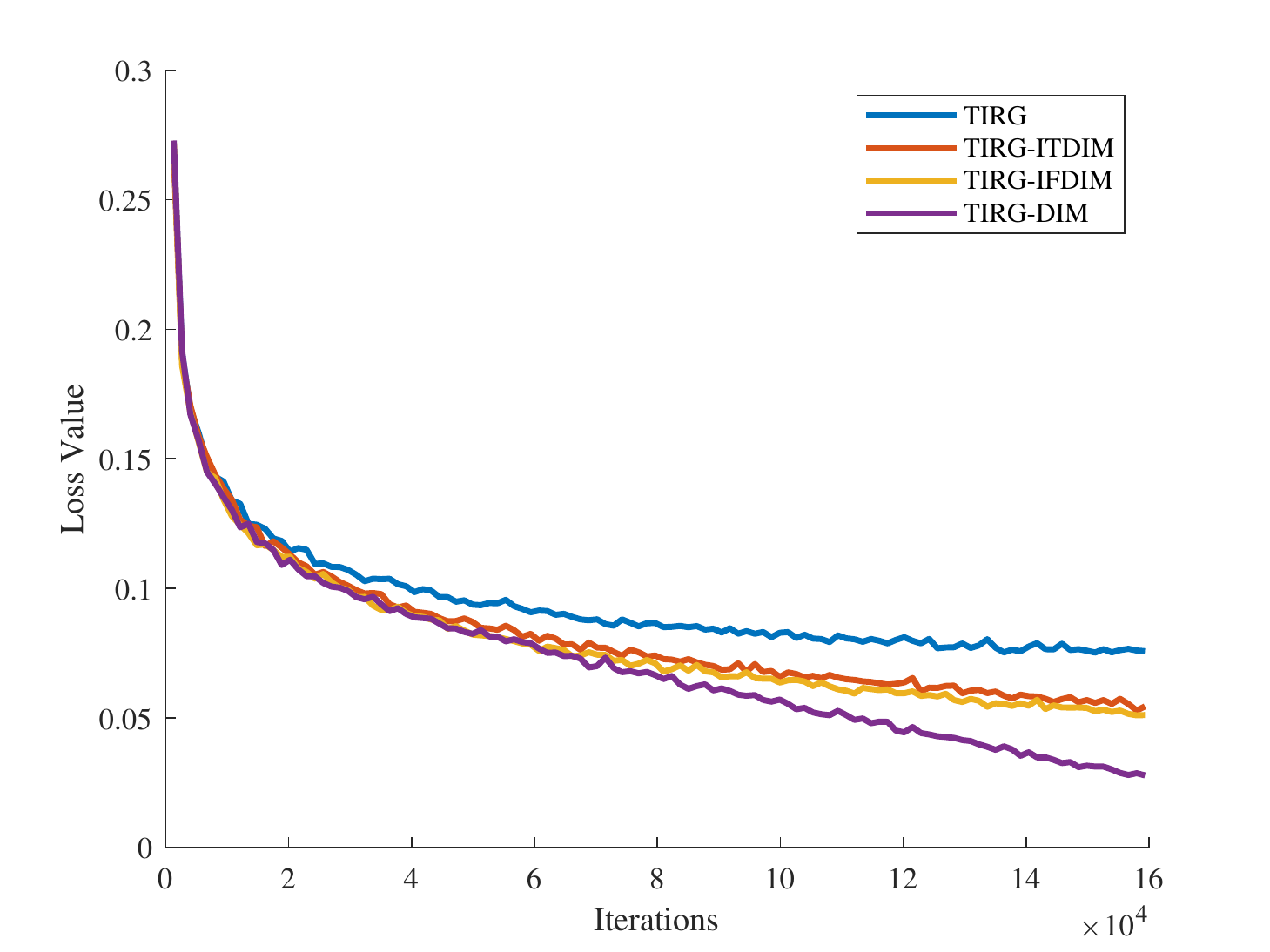}
    \caption{The training loss of the distance metric learning using different models on  MIT-States dataset.}
    \label{fig6}
\end{figure}

Apart from the quantitative analysis of the DIM effect using retrieval accuracy, we also provide qualitative analysis based on the trend of the loss and the visualization of distribution. From Figure \ref{fig6}, we can find that the models' loss trends using different DIMs vary each other. Overall, the models with DIM perform better than that without DIM. The larger the number of iterations, the greater the difference becomes. In detail, TIRG-DIM is superior to the methods with ITDIM or IFDIM. It shows that aligning the distribution of all three modalities performs better than aligning two of them, which is in line with common sense. The TIRG-IFDIM is a little bit better than TIRG-ITDIM in terms of the training loss. This is because the semantic information of the inputs in IFDIM is rich and identical, while the text modality in ITDIM is semantically deficient.

\begin{figure}[htbp]
    \subfigure[The cross-modal invariance preserved without DIM]{
        \includegraphics[width = 0.5\textwidth]{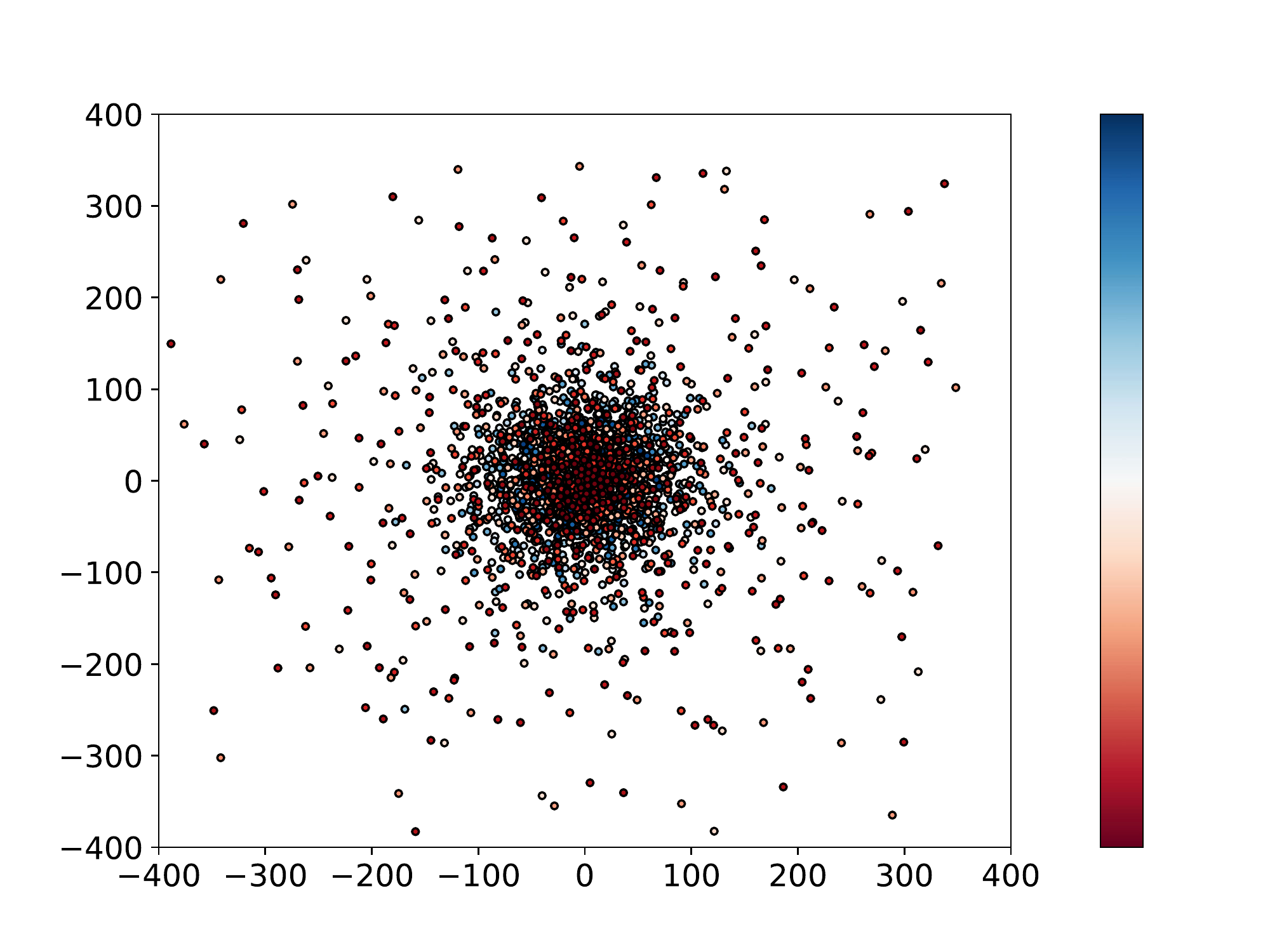}
        \label{fig7:a}
    }
    \subfigure[The cross-modal invariance preserved with ITDIM]{
        \includegraphics[width=0.5\textwidth]{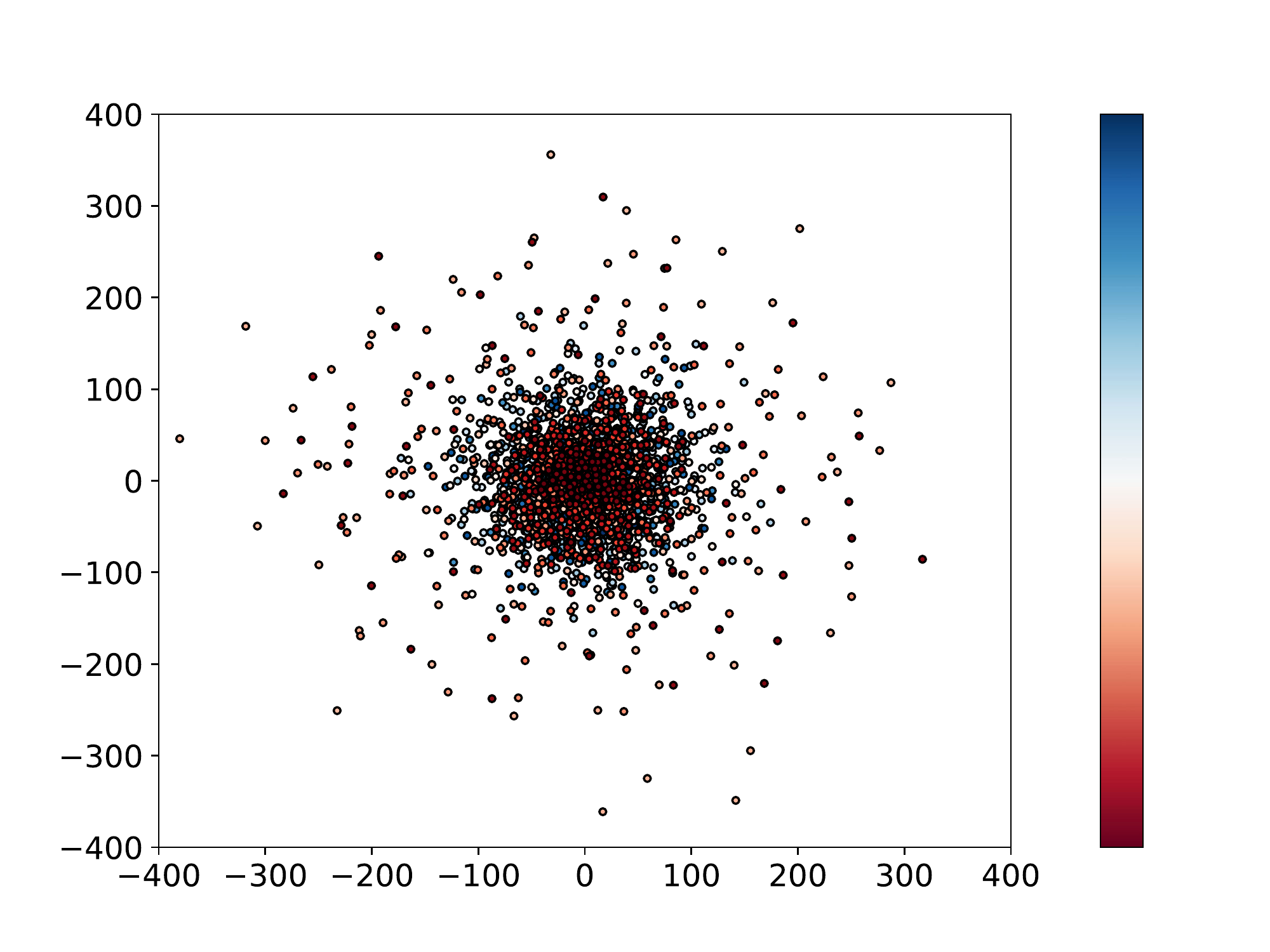}
        \label{fig7:b}
    }
    \subfigure[The cross-modal invariance preserved with IFDIM]{
        \includegraphics[width=0.5\textwidth]{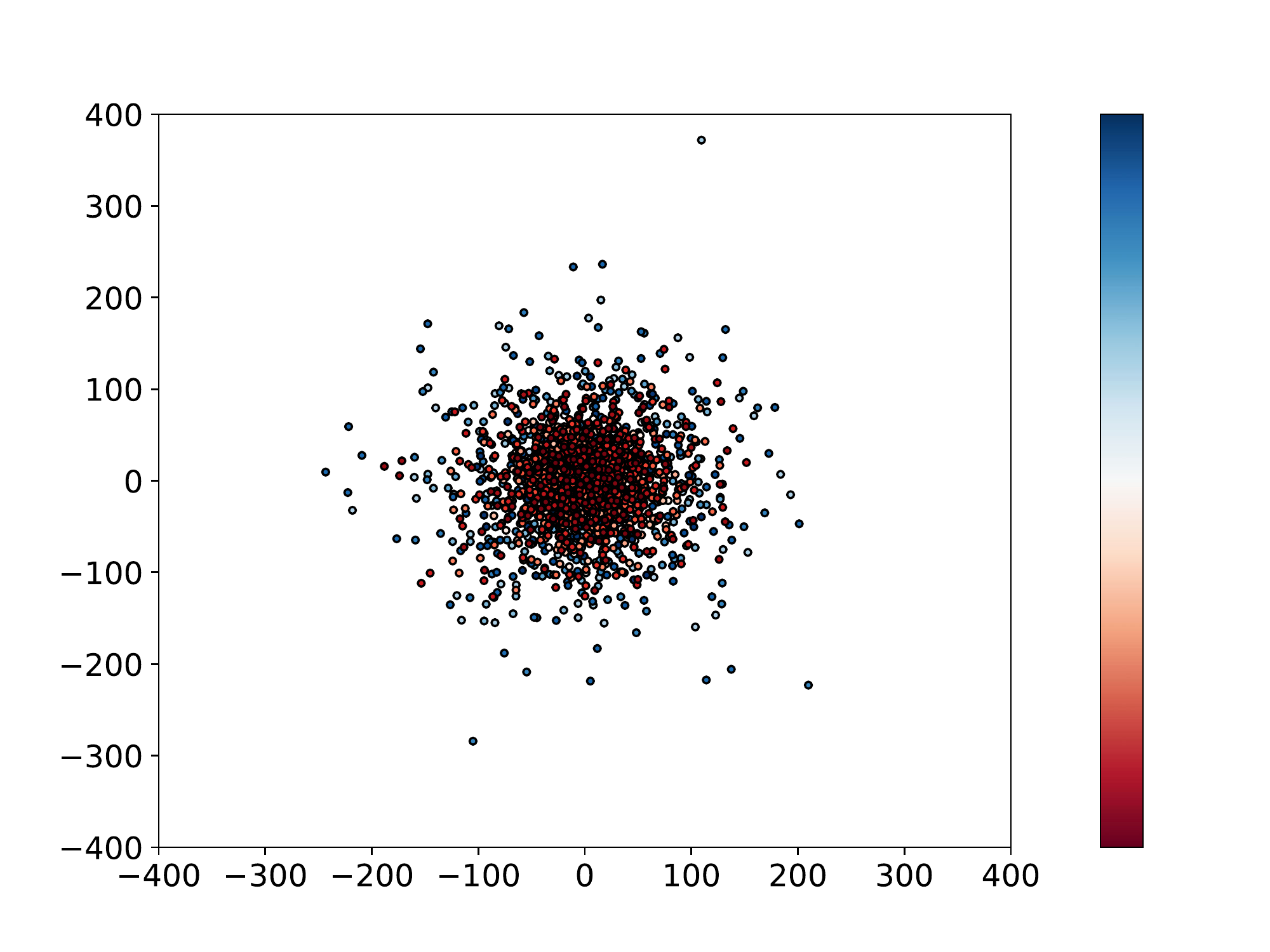}
        \label{fig7:c}
    }
    \subfigure[The cross-modal invariance preserved with DIM]{
        \includegraphics[width=0.5\textwidth]{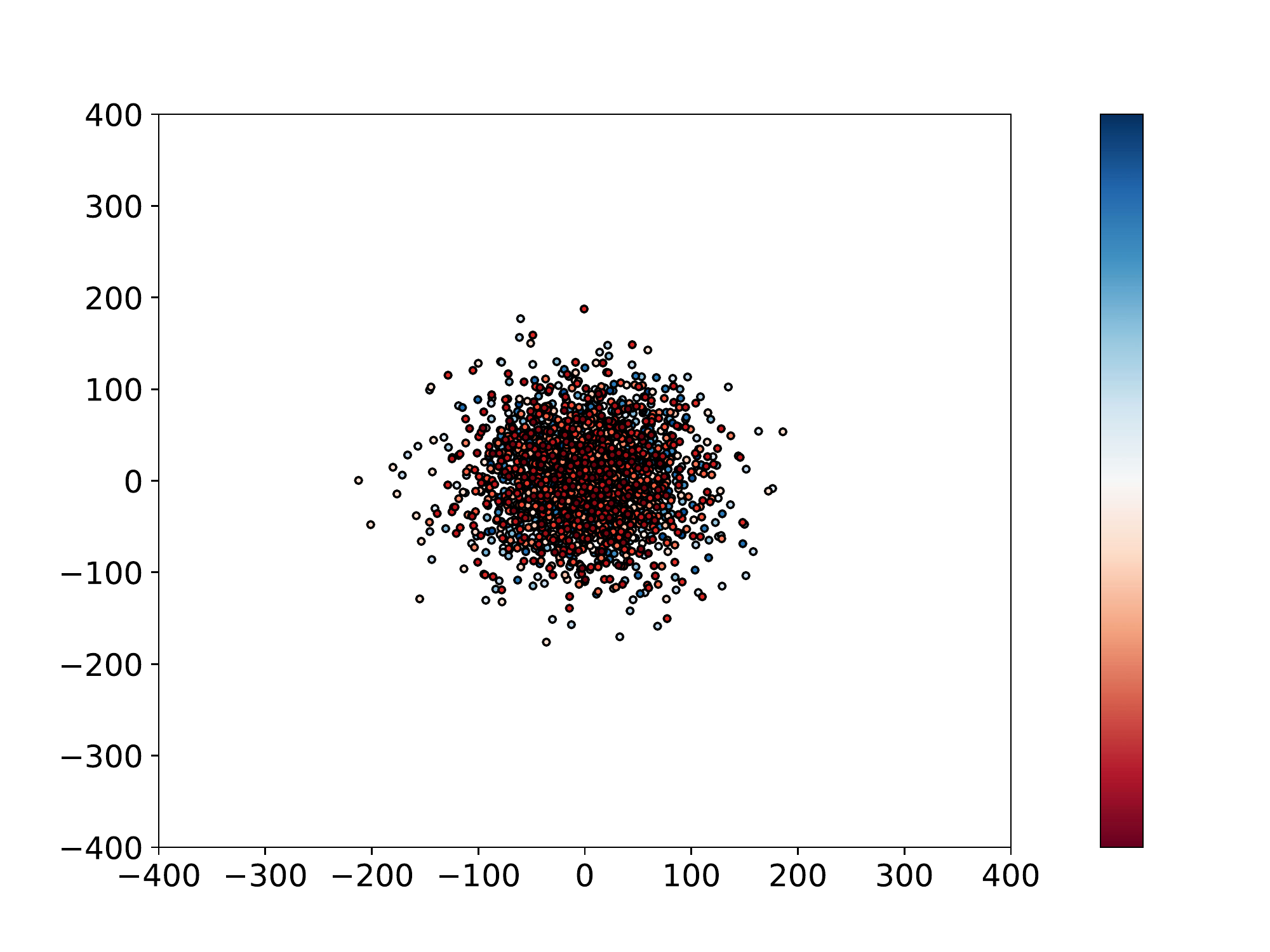}
        \label{fig7:d}
    }
    \caption{The cross-modal invariance preserved. We visualise the distribution of the fusion modality and the image modality using the dots of different colors. The red and green solid dots represent the sampled fusion features and desired image features, respectively. We specify different lightness for the dots of the same color (as shown in the colorbar) to make them easier to distinguish.}
    \label{fig7}
\end{figure}

We also visualize the distributions of the features of the desired image and the fusion of source image and the text on the MIT-states dataset using t-SNE tool (1280 sample points for each modality) by Figure \ref{fig7}. An ideal distribution alignment across modalities should make their features project into a compact common subspace like Figure \ref{fig7:d}, and vice versa. Figure \ref{fig7:a}, Figure \ref{fig7:b}, Figure \ref{fig7:c} and Figure \ref{fig7:d} represent the distribution of the features learned by TIRG, TIRG-ITDIM, TIRG-IFDIM and TIRG-DIM (ITDIM+IFDIM), respectively. The alignment of the distributions of these two features realized by the models with mutual information maximization is much better than that realized by the models without mutual information maximization. Obviously, TIRG-DIM reaches the best alignment of distributions of item representations across modalities. It should be noted that the model based on IFDIM learns a better common subspace for the image modality and the fusion modality compared to the model based on ITDIM, which corresponds to trends of the loss in Figure \ref{fig6}.



\section{Conclusion}
In this paper, we have proposed a new method for cross-modal image retrieval based on the contrastive self-supervised learning method Deep InfoMax \cite{belghazi2018mine,hjelm2018learning}. Our approach makes retrieval more accurate by aligning the feature distributions of text, image, and their fusion. We maximize the MI between semantically different representations of the image modality and the text modality to project the features of these two modalities into a common subspace. Moreover, our method gets a precise alignment of distribution of the image modality and the fusion modality by maximizing the MI between the semantically identical representations in the desired image encoder and the fusion network. The proposed method gives an improved performance on three benchmark datasets. In the future, we would like to apply our work to other areas of cross-modal retrieval.

\section{Acknowledgments}
This work is supported by Alibaba-Zhejiang University Joint Institute of Frontier Technologies, The National Key R$\&$D Program of China (No. 2018YFC2002603, 2018YFB1403202), Zhejiang Provincial Natural Science Foundation of China (No. LZ13F020001), the National Natural Science Foundation of China (No. 61972349, 61173185, 61173186) and the National Key Technology R$\&$D Program of China (No. 2012BAI34B01, 2014BAK15B02).
\bibliography{mybibfile}
\bibliographystyle{elsarticle-num}

\end{document}